\documentclass[letterpaper, 10 pt, conference]{ieeeconf}  
\usepackage{arydshln}

\IEEEoverridecommandlockouts                              

\usepackage{graphics} 
\usepackage{epsfig} 
\usepackage{mathptmx} 
\usepackage{times} 
\usepackage{amsmath} 
\usepackage{amssymb}  
\usepackage{cite}
\usepackage{amsfonts}
\usepackage{algorithmic}
\usepackage{graphicx}
\usepackage{textcomp}
\usepackage{xcolor}
\usepackage{times} 
\usepackage{multirow}
\usepackage{multicol}
\usepackage{url}
\usepackage{afterpage}
\usepackage{stfloats}
\usepackage{arydshln}
\usepackage{epsfig} 
\usepackage{mathptmx}
\usepackage{subcaption}
\usepackage{tablefootnote}
\usepackage{colortbl}  
\usepackage{xcolor}    
\definecolor{lightblue}{rgb}{0.85, 0.92, 1.0}  
\usepackage{arydshln}  
\usepackage{hyperref}
\usepackage{makecell}

\title{\LARGE \bf
MoRAL: Motion-aware Multi-Frame 4D Radar and LiDAR Fusion for Robust 3D Object Detection
}

\author{Xiangyuan Peng $^{1,2\dagger}$  \hspace{2mm}Yu Wang$^{1,2\dagger}$ \hspace{2mm}    Miao Tang$^{3}$ \hspace{2mm}  Bierzynski Kay$^{2}$ \hspace{2mm} Lorenzo Servadei$^{1}$    \hspace{2mm}  Robert Wille$^{1}$
\thanks{$^{\dagger}$Xiangyuan Peng and Yu Wang contribute equally to this work.}
\thanks{$^{1}$Technical University of Munich, Munich, Germany}%
\thanks{$^{2}$Infineon Technologies AG, Neubiberg,
        Germany}%
\thanks{$^{3}$China University of Geosciences, Wuhan, China}%
}

\begin{document}

\maketitle
\thispagestyle{empty}
\pagestyle{empty}


\begin{abstract}
Reliable autonomous driving systems require accurate detection of traffic participants. To this end, multi-modal fusion has emerged as an effective strategy. In particular, 4D radar and LiDAR fusion methods based on multi-frame radar point clouds have demonstrated the effectiveness in bridging the point density gap. However, they often neglect radar point clouds' inter-frame misalignment caused by object movement during accumulation and do not fully exploit the object dynamic information from 4D radar. In this paper, we propose MoRAL, a motion-aware multi-frame 4D radar and LiDAR fusion framework for robust 3D object detection. First, a Motion-aware Radar Encoder (MRE) is designed to compensate for inter-frame radar misalignment from moving objects. Later, a Motion Attention Gated Fusion (MAGF) module integrate radar motion features to guide LiDAR features to focus on dynamic foreground objects. Extensive evaluations on the View-of-Delft (VoD) dataset demonstrate that MoRAL outperforms existing methods, achieving the highest mAP of 73.30\%  in the entire area and 88.68\%  in the driving corridor. Notably, our method also achieves the best AP of 69.67\% for pedestrians in the entire area and 96.25\% for cyclists in the driving corridor. 
\end{abstract}

\section{Introduction}
Modern intelligent transportation system (ITS) relies on robust perception. Various sensors have been applied in autonomous driving for ITS, such as cameras, LiDAR, and radar.
Cameras have been widely used for road perception due to the advanced RGB image algorithms, but suffer from a lack of depth information \cite{mei2024comprehensive}. In contrast, LiDAR sensors provide detailed 3D point clouds. However, LiDAR is sensitive to adverse environments. Small particles from rain, fog, and snow can reduce the signal and cause clutter to LiDAR data \cite{granado2024navigating}. Therefore, LiDAR-only strategies are not enough for practical sensing applications. To compensate for the disadvantages of camera and LiDAR, more research has been developed on radar-based perception \cite{lin2024rcbevdet,bi2025maff,peng2024mufasa}. Especially, 4D radar has gained increasing attention since it generates 3D point clouds with elevation dimensions. Besides, it remains robust under adverse weather and offers velocity and Radar Cross Section (RCS) measurement. However, the point clouds from the 4D radar remain noisy and sparse.

Therefore, some methods fuse 4D radar and LiDAR point clouds for better spatial information and robustness \cite{chae2024towards,huang2024v2x,tiezhen2025bsm}. To bridge the point density gap between the two modalities, a common strategy is to accumulate multi-frame sequential 4D radar data. L4DR \cite{huang2025l4dr} implements bidirectional early fusion of 4D radar and LiDAR point clouds. MutualForce \cite{mutualforce} enhances the representations of both 4D radar and LiDAR at the pillar level through mutual interaction. And RLNet \cite{xu2024rlnet} adaptively weighs the importance of 4D radar and LiDAR features and applies stochastic dropout to mitigate degradation caused by the sensor failure.

\begin{figure}[t]
    \centering
    \includegraphics[width=0.50\textwidth]{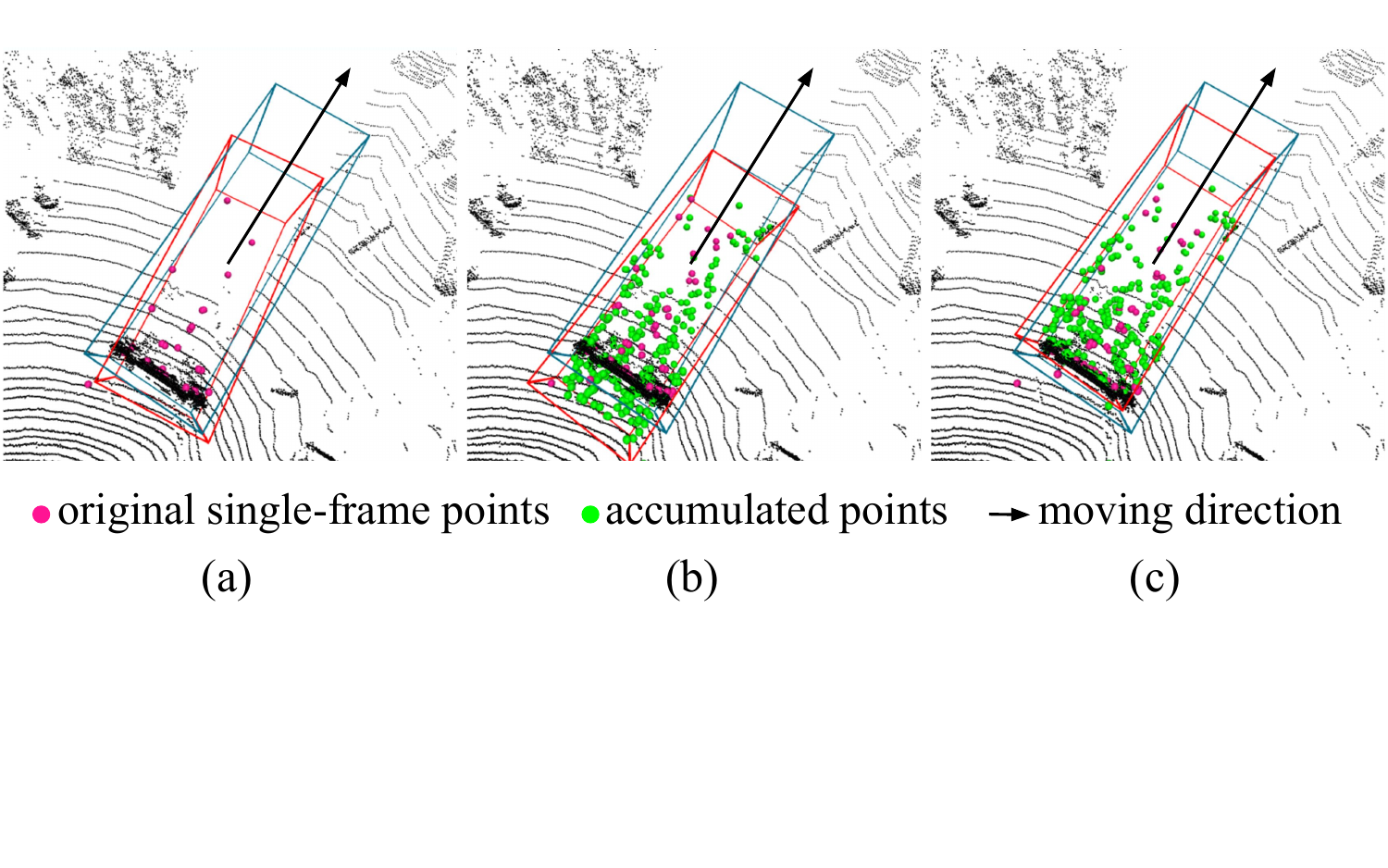} %
    \vspace{-21mm}
    \caption{Visualization of the "tail" issue. Ground truth boxes are blue, and predictions are red. (a) shows a single-frame 4D radar point cloud from VoD dataset \cite{palffy2022multi}. (b) are accumulated multi-frame 4D radar points without motion-aware compensation. (c) denotes accumulated 4D radar point clouds with our MoRAL.}
    \label{fig:motivation}
    \vspace{-4mm}
\end{figure}

However, these methods only considered ego-motion compensation on multi-frame 4D radar data. The misalignment from the movement of dynamic objects during temporal accumulation is neglected. As a result, although 4D radar data achieves higher point density, points from different frames belonging to the fast-moving objects can be shifted. As shown in Fig. \ref{fig:motivation}(b), we observed that directly stacking multiple frames of 4D radar point clouds causes moving objects to be stretched along their motion direction, generating a "tail" in final point clouds. This "tail", which is commonly found in multi-frame LiDAR accumulation \cite{chen2022mppnet}, also exists in multi-frame 4D radar point clouds. Especially, due to the sparsity of 4D radar point clouds, the impact of the "tail" becomes more severe, leading to potential false positives and shape distortion.

To address these challenges, we introduce MoRAL, a 4D radar and LiDAR fusion framework that addresses the misalignment of dynamic objects in multi-frame 4D radar point clouds through object motion compensation. The radar motion features are extracted via a Moving Object Segmentation (MOS)-based encoder to accurately infer object motion status and conduct corresponding point-level motion compensation. 
Besides, the radar motion features are used to enhance the LiDAR spatial features, reducing foreground-background confusion in LiDAR representations.
The main contributions are as follows:
\begin{itemize}
\item We address the objects' inter-frame misalignment from 4D radar accumulation and propose a Motion-Aware Radar Encoder (MRE) to mitigate the motion-induced noise while enhancing point density. 
\item A Motion Attention Gated Fusion (MAGF) module incorporates radar motion information into the LiDAR branch, guiding LiDAR features to focus on dynamic foreground objects while neglecting background clutter. 
\item Extensive experiments on View-of-Delft (VoD) dataset \cite{palffy2022multi} demonstrate the effectiveness of our method. 
\end{itemize}

\section{Related Work}

\subsection{Single-modal 3D Object Detection}
Camera-based methods are widely applied due to the cost efficiency and rich semantic information \cite{detr3d,ocbev,graphdetr4d}. However, the absence of accurate depth information limits their applications.

LiDAR-based detection, on the other hand, provides precise geometric measurements. Grid-wise methods \cite{voxelnet,pointpillars,swiftpillars} discretize LiDAR point clouds into voxels or pillars and apply 2D convolutions for fast feature extraction, inevitably leading to information loss. To address this issue, point-wise methods \cite{liu2024pp,li2024ws,song2023psns}  operate directly on raw point clouds to preserve fine-grained geometric details. Additionally, hybrid approaches \cite{chen2024point, shi2023pv, jiang2024dpa} utilize both representations for a better trade-off between computation cost and information loss.

However, the robustness of LiDAR significantly degrades under adverse weather conditions. In contrast, 4D radar offers all-weather robustness and provides additional Doppler velocity and RCS information. RadarPillars \cite{radarpillars} pillarizes 4D radar point clouds and enhances radar features by decomposing absolute radial velocity. MAFF-Net \cite{bi2025maff} leverages radial velocity for point clustering. MVFAN \cite{yan2023mvfan} and MUFASA \cite{peng2024mufasa} utilize both cylindrical and Bird's Eye View (BEV) perspectives for improved spatial awareness.

\subsection{Multi-modal 3D Object Detection}
Although single-modal methods have been extensively developed \cite{trigka2025comprehensive}, they remain constrained by the inherent weaknesses of individual sensors, such as point sparsity. Therefore, multi-modal methods have been explored.

LiDAR and camera fusion \cite{fan2024snow,gafusion,sparselif} benefits from the complementarity between geometry and semantic information. Meanwhile, radar and camera fusion addresses the challenges of poor lighting conditions and adverse weather. RCFusion \cite{zheng2023rcfusion} projects 4D radar and image features into a unified BEV space. RobuRCDet \cite{roburcdet} dynamically fuses 4D radar features with image features, guided by image confidence scores associated with different weather conditions.

Compared to vision-based fusion methods, 4D radar and LiDAR fusion methods enable accurate spatial geometry with all-weather robustness, making it particularly suitable for dynamic perception. InterFusion \cite{interfusion} and M$^2$Fusion \cite{wang2022multi} use attention mechanisms to fuse pillarized 4D radar and LiDAR data. L4DR \cite{huang2025l4dr} denoises LiDAR data with 4D radar point clouds through diffusion, and MutualForce \cite{mutualforce} exploits radar-specific features like velocity and RCS to guide the fusion process. Although these methods all use accumulated multi-frame 4D radar point clouds, they overlook the inter-frame dynamic object misalignment in 4D radar point clouds. 
Our proposed MoRAL addresses this issue through MOS-based motion compensation.

\subsection{Moving Object Segmentation}
MOS is applied to differentiate dynamic objects from static backgrounds.\cite{chen2021moving} segments object motion by transforming sequential LiDAR scans into range images and deriving residual maps. Compared to LiDAR, 4D radar provides velocity information, enabling a single frame segmentation without sequential object tracking. Radar Velocity Transformer \cite{zellerradarvelo} enhances MOS performance by employing attention mechanisms. And RadarMOSEVE \cite{radarmoseve} leverages relative radial velocity for joint MOS and ego-velocity estimation task.

Dynamic objects are essential for autonomous driving. However, the motion status of objects has not been effectively utilized to enhance detection tasks. Thus, our approach integrates motion information into a 4D radar and LiDAR fusion framework to achieve better 3D object detection.

\section{Proposed Method}
\subsection{Overall Structure}
\begin{figure*}[t]
    \centering
    \includegraphics[width=0.95\textwidth]{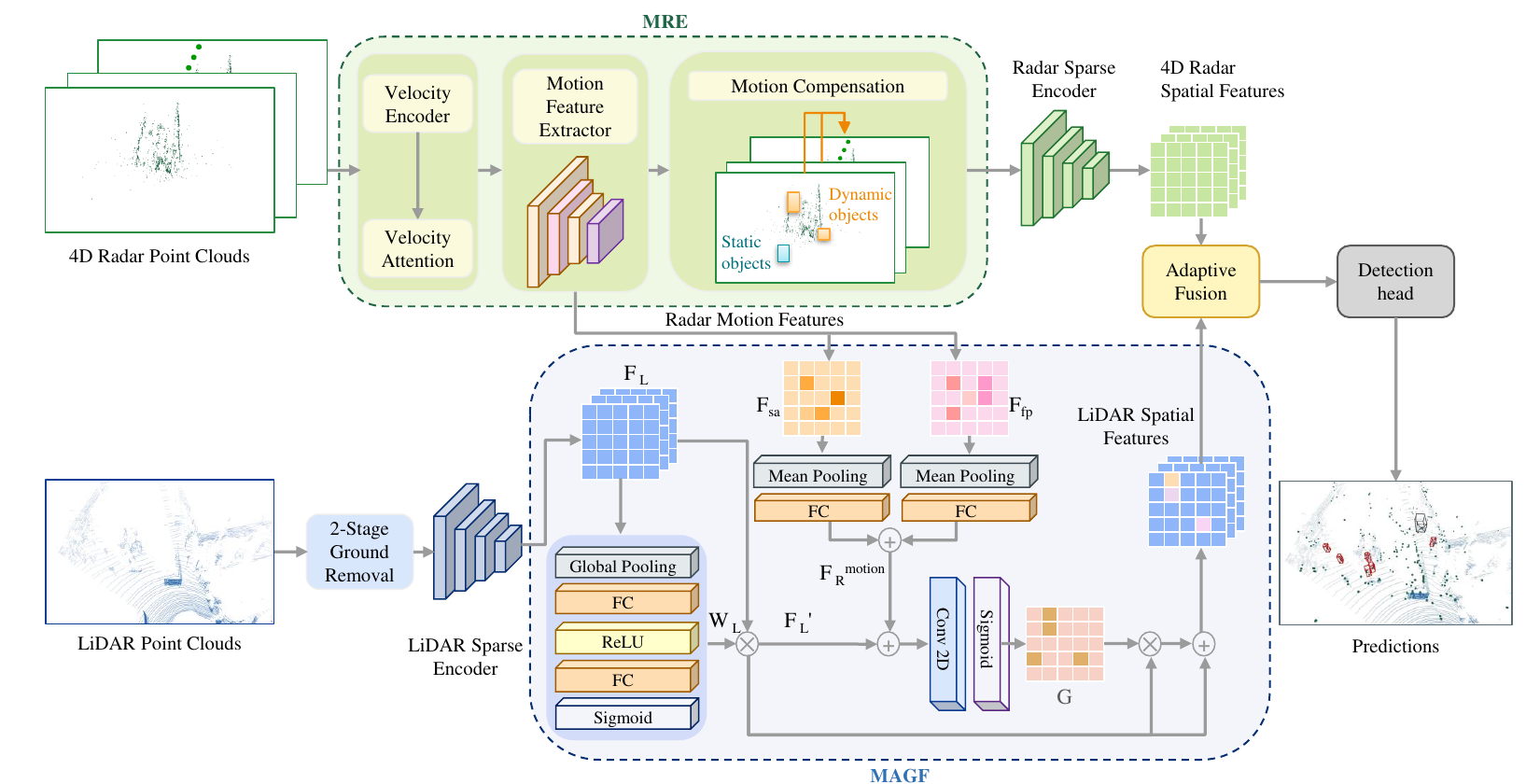} %
    \caption{The overall structure of our MoRAL.}
    \label{fig: overall structure}
    \vspace{-0mm}
\end{figure*}

This section presents the structure of our proposed MoRAL. The overall architecture is shown in Fig. \ref{fig: overall structure}. First, the multi-frame 4D radar point clouds are processed by the MRE module, generating radar motion features and motion-compensated 4D radar point clouds. The compensated 4D radar point clouds are then used to extract 4D radar spatial features through a radar sparse encoder. 
In parallel, single-frame LiDAR point clouds are fed into a  two-stage RANSAC-based \cite{ransac} filter for ground points removal and extracted through a LiDAR sparse encoder to obtain LiDAR spatial features. The LiDAR spatial features are subsequently enhanced by radar motion features through the MAGF module, and further fused with radar spatial features through the Adaptive Fusion from RLNet \cite{xu2024rlnet}. Finally, the detection head will predict 3D bounding boxes.

\subsection{Motion-aware Radar Encoder}
\begin{figure}[t]
  \centering
  \hspace*{-2.5mm}
  \begin{subfigure}[b]{0.22\textwidth} 
    \fbox{\includegraphics[width=\textwidth]{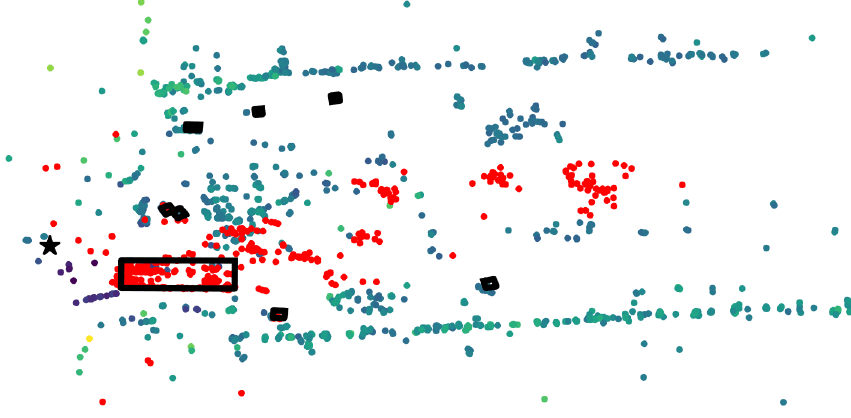}}
    \caption{}
  \end{subfigure}
  \hspace{2mm}
  \begin{subfigure}[b]{0.22\textwidth}
    \fbox{\includegraphics[width=\textwidth]{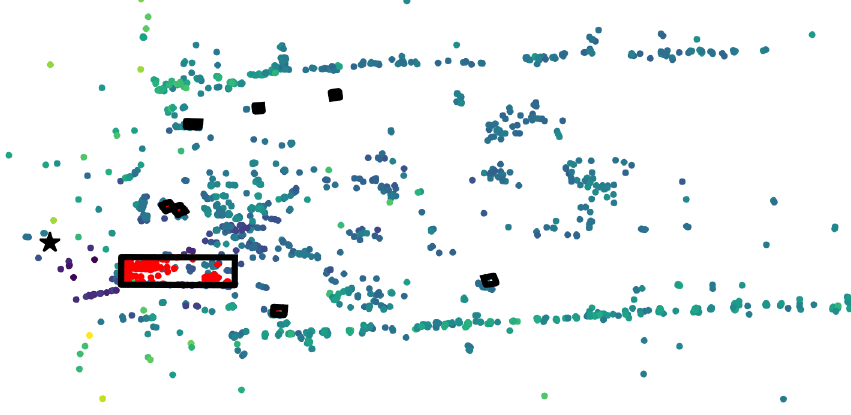}}
    \caption{}
  \end{subfigure}
  \vspace{-1mm}
  \caption{Illustration of segmentation results. (a) shows the velocity threshold-based segmentation, while (b) presents the MOS ground truth. Red and green denote moving and static points. The color intensity of static points reflects the RCS value. Black boxes are ground truth bounding boxes.
}
  \label{fig:seg_comp}
  \vspace{-3mm}
\end{figure}

Temporal accumulation across multiple frames is widely applied to overcome the sparsity of 4D radar point clouds \cite{xu2024rlnet,huang2025l4dr,tan20223}. However, current methods \cite{xu2024rlnet,tan20223} only account for radar ego-motion compensation to obtain the absolute radial velocities. The inter-frame misaligned "tail" caused by objects' motion during accumulation, as shown in Fig. \ref{fig:motivation}(b), is often overlooked, leading to inaccurate detections.
To eliminate the "tail" of 4D radar point clouds, we consider motion compensation for moving objects, which requires segmenting objects' motion status. Due to the velocity information provided by 4D radar, segmentation can be more intuitively achieved without cross-frame tracking \cite{chen2021moving}. However, direct segmentation via velocity threshold suffers from noise and the multi-path effect. As shown in Fig. \ref {fig:seg_comp}(a), the $1m/s$ absolute velocity threshold-based segmentation misclassifies a large number of static background points as moving. Therefore, we propose the MOS-based MRE module to accurately infer the motion status and perform point-level motion compensation. 

\begin{figure}[t]
    \centering
\includegraphics[width=0.48\textwidth]{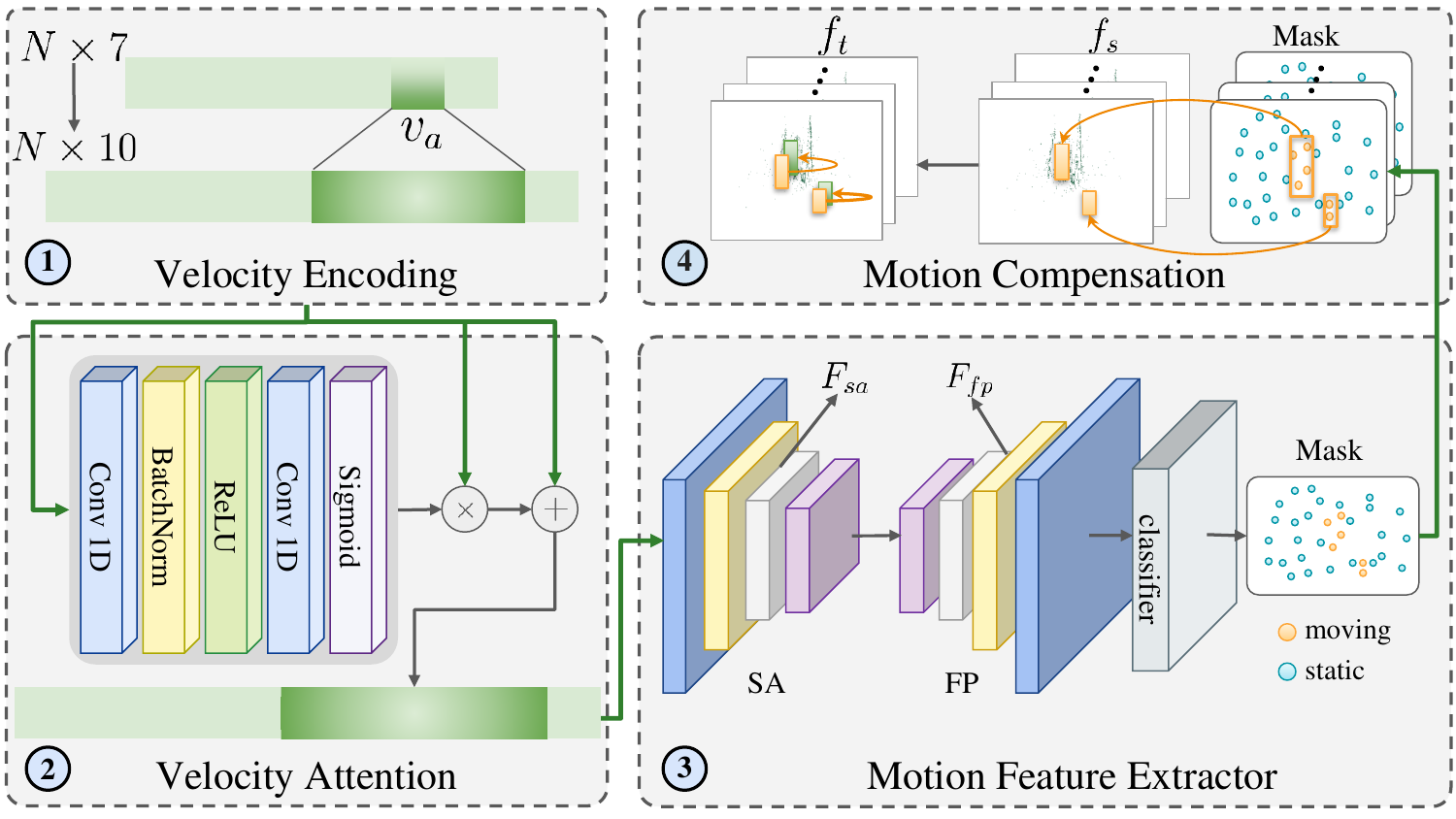} %
    \vspace{-3mm}
    \caption{Architecture of MRE module.}
    \label{fig:mos}
    \vspace{-7mm}
\end{figure}

As illustrated in Fig. \ref{fig:mos}, MRE comprises four components: Velocity Encoding, Velocity Attention, Motion Feature Extractor, and Motion Compensation.
Given an accumulated 4D radar point cloud $P\in\mathbb{R}^{ N\times 7}$,  where $N$ denotes the number of radar points and each point comprises seven features: location (x, y, z), RCS, relative and absolute radial velocity, and timestamp. The absolute radial velocity $v_a$ is first augmented in Velocity Encoding by computing its magnitude, squared value, and moving direction. These quantities are concatenated with the original features to form a velocity-enhanced point cloud $P_{enhanced}\in\mathbb{R}^{ N\times10}$. $P_{enhanced}$ is subsequently processed by the Velocity Attention, which computes point-wise attention weights to selectively enhance $v_a$. The Velocity Attention highlights dynamic points and provides a more discriminative representation for downstream segmentation. The resulting features are then passed to the Motion Feature Extractor, comprising three Set Abstraction (SA) layers, three Feature Propagation (FP) layers, and a point-wise classifier, to learn hierarchical motion information and predict motion status. For training, MOS labels are generated based on object-level motion annotations. The predicted motion label $\tilde{y}_i \in \{0,1\}, i=0,\dots, N-1$  for each point $p_i$ is determined by a motion parameter $\alpha$, where $0$ and $1$ indicate static and moving points, respectively.

Point-level motion compensation is subsequently applied. Let $f_s$  and \( f_{\text{t}} \) denote the source and target frame in accumulation. We first define a motion mask $M_{}^{pred} \in \{0,1\}^N$, where $M_{i}^{pred}=\tilde{y}_i,$ indicating whether point $p_i^{f_s} \in P_{enhanced}$  from source frame is identified as moving or static. Then for each point $p_i^{f_s}$, the absolute radial velocity vector $\mathbf{v}^{f_s}_{a,i}$ relative to the radar origin is calculated by:
\begin{equation}
\mathbf{u}_i^{f_s}=\frac{(x_i^{f_s},y_i^{f_s},z_i^{f_s})}{\sqrt{(x_i^{f_s})^2 + (y_i^{f_s})^2 + (z_i^{f_s})^2}},
\end{equation}
\begin{equation}
    \mathbf{v}^{f_s}_{a,i} = M_{i}^{pred}\cdot [v^{f_s}_{a,i}\cdot \mathbf{u}_i^{f_s}],
\end{equation}
where $\mathbf{u}_i^{f_s}$ represents absolute radial unit vector. For each point $p_i^{f_s}$ in source frame ${f_s}$,  its motion-compensated position  $\tilde{p}_i^{f_t}$ in target frame $f_{t}$  is computed as:
\begin{equation}
    \tilde{p}_i^{f_{t}}=p_i^{f_s} + M^{pred}_{i}\cdot[\tau\cdot (f_{t}-{f_s})\cdot \mathbf{v}_{a,i}^{f_s}],
\end{equation}
where $\tau$ represents the sampling frequency of the radar sensor, $ M_{i}^{pred}$ ensures that only points predicted as moving are compensated, while static points remain unchanged. Fig. \ref{fig:motivation}(c) illustrates our compensation results.

\subsection{Motion Attention Gated Fusion}
Radar motion features, $F_{sa}$ and $F_{fp}$ from the Motion Feature Extractor, captured enriched dynamic information. We adaptively incorporate these motion features to highlight LiDAR foreground features and better distinguish them from the noisy background using the MAGF module. 

As shown in Fig. \ref{fig: overall structure}, we first apply a channel attention mechanism to LiDAR spatial features $F_L$, which adaptively adjusts channel-wise importance. $F_L$ undergoes a global average pooling $P_g$, two fully connected layers $FC$ with a ReLU activation in between, followed by a Sigmoid function $\sigma$ to generate channel-wise attention weights $W_L$. $W_L$ are further applied to enhance $F_L$ and obtain the recalibrated $F_L^{'}$ as follows:
\begin{equation} W_L = \sigma\left(\text{FC}_2\left(\text{ReLU}\left(\text{FC}_1\left(\text{P}_{g}(F_L)\right)\right)\right)\right), \quad F_L^{'} = F_L \odot W_L, \end{equation}
In parallel, radar motion features $F_{sa}$ and $F_{fp}$ are separately processed through mean pooling followed by fully connected projections. A learnable parameter $\lambda \in (0,1)$ is introduced to adaptively balance the global motion information from $F_{sa}$ and local motion information from $F_{fp}$. The resulting multi-scale motion features are aggregated via weighted summation and reshaped to form the unified motion feature $F_R^{motion}$ as follows:
\begin{equation}
F_R^{motion}=\psi(\lambda \cdot  \phi_{sa}(F_{sa}) + (1-\lambda)\cdot\phi_{fp}(F_{fp})),
\end{equation}
where $\psi(\cdot)$ denotes the reshaping and $\phi(\cdot)$ represents channel-wise mean pooling and fully connected projections. Compared to direct concatenation, the adaptive aggregation mitigates local noise and redundancy in radar motion features, resulting in a more compact and reliable motion representation $F_R^{motion}$.

Furthermore, a gating map $G$ is calculated by fusing $F_L^{'}$ and $F_R^{motion}$:
\begin{equation}
G=\sigma(\text{Conv}(\text{Concat}(F_L^{'},F_R^{motion}))),
\end{equation}
The gating map guides the LiDAR features enhancement and obtain final LiDAR spatial features as:
\begin{equation}
F^{enhanced}_L=F_L^{'}\odot G+F_L^{'}.
\end{equation}
By incorporating radar motion features, MAGF augments LiDAR features with object motion awareness while suppressing irrelevant information. Unlike fusion methods \cite{xu2024rlnet,lirafusion} that rely solely on attention-weighted feature allocation, MAGF integrates object motion cues into LiDAR representations, leading to stronger feature enhancement for dynamic foreground areas.

\section{Experiments}
\label{Experiments}

In this section, we compare our MoRAL with existing 3D object detection methods. All models are trained for 80 epochs with a batch size of 8 using a single NVIDIA RTX 4070 GPU. We adopt the Adam optimizer \cite{kingma2014adam} with an initial learning rate of 0.003 and a weight decay of 0.01. To improve model robustness and generalization, we apply standard data augmentation techniques including flipping, scaling, and rotation.
The implementation is built upon the OpenPCDet \cite{od2020openpcdet}, a widely used library for 3D point cloud. 

\subsection{Dataset and Metrics}
The proposed MoRAL is evaluated on the VoD dataset \cite{palffy2022multi}, since it provides object-level motion status labels. It contains 8,693 frames of synchronized 4D radar, LiDAR, and camera data, primarily collected in urban scenes with diverse traffic participants. As the official test server is unavailable, all evaluations are conducted on the validation set \cite{10268601}.

To evaluate our method, we report per-class Average Precision (AP) and mean Average Precision (mAP) across all categories. An IoU threshold of 50\% is used for cars, while a lower threshold of 25\% is applied for pedestrians and cyclists. Consistent with the evaluation protocol in the original paper \cite{palffy2022multi}, we assess detection performance within two regions: the entire area and the driving corridor.

\subsection{Main Results}
We compare our model with current single- and multi-modal methods in Table \ref{tab: main result vod}. From Table \ref{tab: main result vod}, our MoRAL achieves the best mAP of 73.30\% and 88.68\% in the entire area and driving corridor, respectively. Notably, for pedestrians, our approach achieves the best AP with 69.67\% in the entire area. For cyclists, our method outperforms RLNet \cite{xu2024rlnet} by 4.58\% AP in the driving corridor. 

Our method achieves better performance for pedestrians and cyclists, as the majority of these two categories in the VoD dataset \cite{palffy2022multi} are in motion \cite{mutualforce}. This demonstrates the effectiveness of our MRE module for moving objects.
In contrast, since most cars in the VoD dataset \cite{palffy2022multi} are stationary, the detection enhancement for cars is less pronounced. Besides, the improvement of the driving corridor is higher since fewer background points are moved during motion compensation. Additionally, our model delivers a real-time inference speed of 15.22 FPS.

\begin{figure*}[t]
    \centering
    \includegraphics[width=0.98\textwidth]{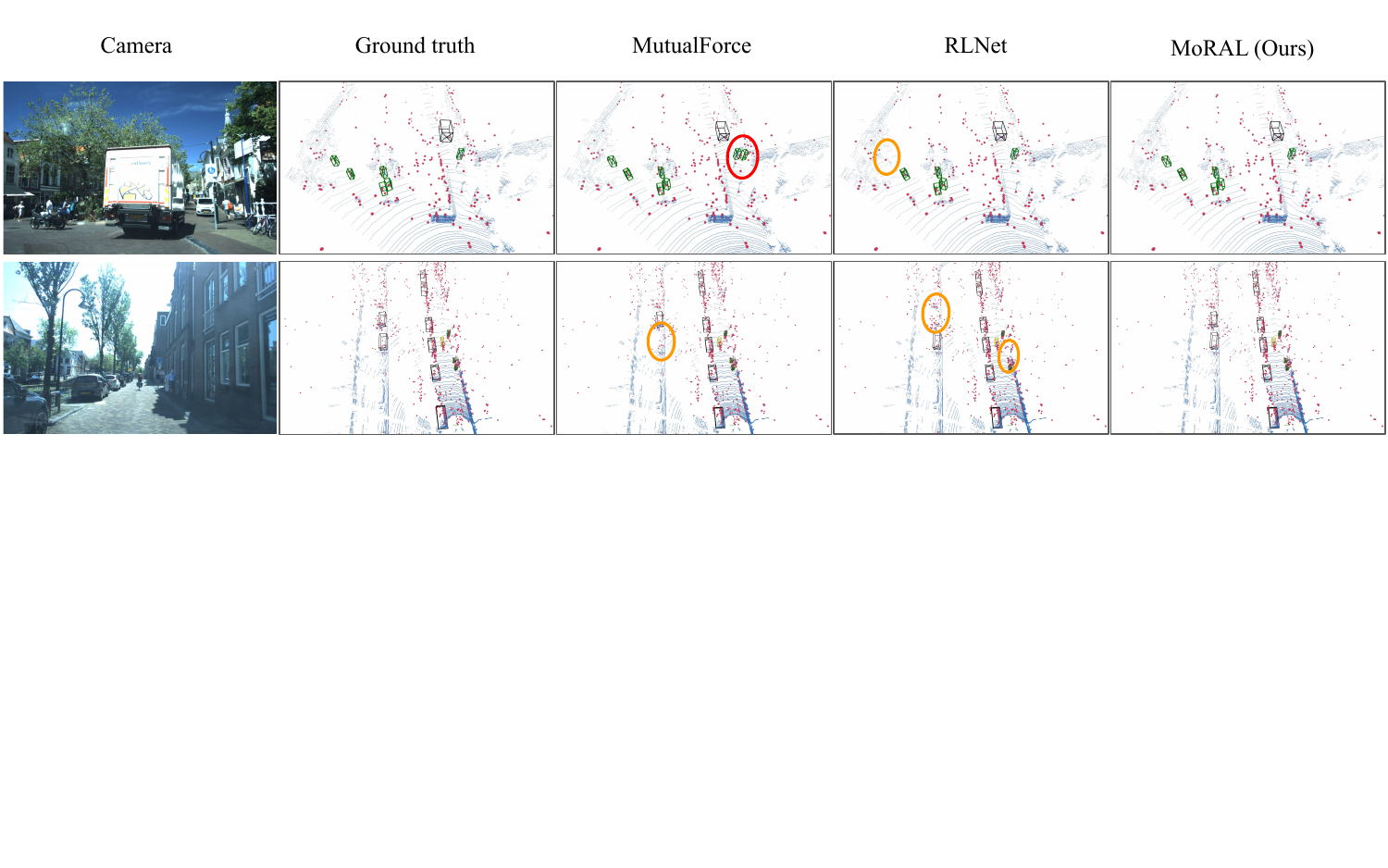} %
    \vspace{-55mm}
    \caption{Qualitative results comparing our method with MutualForce \cite{mutualforce} and RLNet \cite{xu2024rlnet}. Green, yellow, and black boxes denote pedestrians, cyclists, and cars, respectively. Orange and red circles show the false negatives and positives.}
    \label{fig: qualitative}
    \vspace{-5mm}
\end{figure*}

Fig. \ref{fig: qualitative} shows the qualitative results of MutualForce \cite{mutualforce}, RLNet \cite{xu2024rlnet}, and our MoRAL. Compared to the other two methods, our approach demonstrates a more robust detection performance with fewer false negatives and positives.

\setlength\tabcolsep{11pt}
\begin{table*}[ht]
\centering
\caption{Comparative AP (\%) results on VoD val. set \cite{palffy2022multi}. The best results are bold, and the second best are underlined.}
\label{tab: main result vod}
\begin{tabular}{c|c|c|cccc|cccc}
\hline
\multirow{2}{*}{Methods} &\multirow{2}{*}{Modality} & \multirow{2}{*}{Year} & \multicolumn{4}{c|}{Entire Area}  & \multicolumn{4}{c}{Driving Corridor}\\
\cline{4-11}
 & & &Car & Ped. &Cyc. & mAP & Car & Ped. & Cyc. &  mAP\\
\hline
PointPillars\cite{lang2019pointpillars} &R & 2019&37.92   &31.24  &65.66  &44.94  &71.41  &42.27  &87.68  &67.12 \\
PV-RCNN$^\dagger$\cite{shi2020pv} &R &2021 &41.65   &38.82  &58.36  &46.28  &72.00  &43.53  &78.32  &64.62 \\
MVFAN$^\dagger$\cite{yan2023mvfan} &R & 2023&38.12 &30.96  &66.17  &45.08  &71.45 &40.21 &86.63 &66.10 \\
SMURF\cite{10274127} &R & 2023&42.31 &39.09  &71.50  &50.97  &71.74  &50.54  &86.87  &69.72 \\
MUFASA$^\dagger$\cite{peng2024mufasa} &R & 2024&43.10  &38.97 &68.65 &50.24 &72.50  &50.28 &88.51  &70.43\\
MAFF-Net$^\dagger$\cite{maff} &R & 2025&42.33  &46.75 &74.72 &54.59 &72.28  &57.81 &87.40  &72.50\\
DADAN \cite{wang2025dadan} &R & 2025&46.82  &45.20 &74.61 &55.54 &79.32  &51.42 &86.29  &72.34\\
\cdashline{1-10}
BEVFusion\cite{liu2023bevfusion}  &R+C &2023 &37.85 &40.96 &68.95 &49.25 &70.21 &45.86  &89.48  &68.52 \\
RCFusion\cite{zheng2023rcfusion}  &R+C & 2023&41.70 &38.95 &68.31 &49.65 &71.87 &47.50  &88.33  &69.23 \\
LXL\cite{xiong2023lxl} &R+C &2023 &42.33 &49.48 &77.12 &56.31 &72.18 &58.30 &88.31 &72.93 \\
RCBEVDet\cite{lin2024rcbevdet} &R+C & 2024&40.63 &38.86  &70.48  &49.99  &72.48  &49.89  &87.01  &69.80\\
SGDet3D\cite{sgdet3d} &R+C &2024 &53.16 &49.98 &76.11 &59.75 &81.13 &60.91 &90.22 &77.42 \\
LXLv2\cite{xiong2025lxlv2} &R+C &2025 &47.81 &49.30 &77.15 &58.09 &- &- &- &- \\
\cdashline{1-10}
PointPillars$^\dagger$ \cite{lang2019pointpillars} &L & 2019&65.55  &55.71  &72.96  &64.74  &81.10  &67.92  &88.96  &79.33 \\
LXL-Pointpillars \cite{xiong2023lxl} &L & 2023&66.60 &56.10 &75.10 &65.90 &-  &-  &-  &- \\
\cdashline{1-10}
InterFusion$^\dagger$\cite{wang2022interfusion} &R+L &2022 &67.50 &63.21  &78.79 &69.83 &88.11  &74.80  &87.50  &83.47\\
MutualForce\cite{mutualforce} &R+L &2024 &\underline{71.67} &66.26 &77.35 &71.76  &\textbf{92.31} &76.79&89.97 &86.36  \\
L4DR\cite{huang2025l4dr} &R+L &2024 &69.10 &66.20 &\textbf{82.80} &72.70  &90.80 &76.10 &\underline{95.50} & \underline{87.47}\\
CM-FA$^\dagger$\cite{deng2024robust} &R+L & 2024&\textbf{71.39} & 68.54  &76.60  &72.18 &90.91  & \textbf{80.78} &87.80  &86.50  \\
RLNet$^\dagger$\cite{xu2024rlnet} & R+L & 2024& 70.88 & \underline{69.43} & 78.12 & \underline{72.81} & 90.82 & 78.71 & 91.67 & 87.07 
\\
\cdashline{1-10}
\rowcolor{lightblue}
MoRAL (Ours) & R+L &2025 & 71.23 & \textbf{69.67} &\underline{79.01} & \textbf{73.30} & \underline{90.91} &\underline{78.90} &\textbf{96.25} &\textbf{88.68}\\
\hline
\end{tabular}
\\[2pt]
\scriptsize{R, C, and L denote the 4D radar, camera, and LiDAR.  $\dagger$ indicates reproduced results.}
\vspace{-4mm}
\end{table*}

\subsection{Ablation Study}
To investigate the impact of key modules on overall detection performance, we conduct extensive ablation studies.

\setlength\tabcolsep{3pt}
\begin{table}[t]
\centering
\caption{Analysis of different modules.}
\vspace{-1mm}
\label{tab:module_analysis}
\begin{tabular}{c|c|cccc|cccc}
\hline
\multirow{2}{*}{MRE} &\multirow{2}{*}{MAGF} & \multicolumn{4}{c|}{All Area}   & \multicolumn{4}{c}{Driving Corridor}\\
\cline{3-10}
 &  &Car & Ped. &Cyc. & mAP & Car & Ped. & Cyc. &  mAP\\
\hline
 & & 70.35 & 68.81 & 77.49 & 72.22 & 90.21 & 77.52 & 87.38 & 85.04 \\
$\checkmark$ & & \textbf{71.36} & 69.21 & 78.66 & 73.08 & 90.89 & 78.19 & 94.01 & 87.70 \\
 & $\checkmark$ & 70.67 & 69.58 & 78.78 & 73.01 & 90.74 & 78.66 & 92.21 & 87.20 \\
\rowcolor{lightblue}
$\checkmark$ & $\checkmark$ & 71.23& \textbf{69.67} & \textbf{79.01} & \textbf{73.30} & \textbf{90.91} & \textbf{78.90} & \textbf{96.25} & \textbf{88.68} \\
\hline
\end{tabular}
\\[2pt]
\vspace{-6mm}
\end{table} 
\textbf{Analysis of different modules:} Experiments with different modules are presented in Table \ref{tab:module_analysis}. First, we exclude both the MRE and MAGF modules and fuse the spatial features extracted from two sparse encoders directly. It is worth noting that the MAGF module is functionally coupled with the motion feature extractor in MRE. According to Table \ref{tab:module_analysis}, using MRE and MAGF modules alone yields mAP improvements of 2.66\% and 2.16\% in the driving corridor. When both the MRE and MAGF modules are employed, the network achieves the best performance, with mAP improvements of 1.08\% in the entire area and 3.64\% in the driving corridor.

\setlength\tabcolsep{3pt}
\begin{table}[t]
\centering
\caption{Radar point cloud density for 4D radar only model (PointPillars) \cite{pointpillars}.}
\vspace{-1mm}
\label{tab:radar_only}
\begin{tabular}{c|cccc|cccc}
\hline
\multirow{2}{*}{Radar Frames} & \multicolumn{4}{c|}{All area} & \multicolumn{4}{c}{Driving Corridor} \\
\cline{2-9}
& Car & Ped. & Cyc. & mAP & Car & Ped. & Cyc. & mAP \\
\hline
1 frame            & 37.53 & 29.05 & 65.79 & 44.12 & 71.30 & 34.95 & 87.58 & 64.61 \\
3 frames       & 38.68 & 29.87 & 66.43 & 44.99 & 71.83 & 36.18 & 88.92 & 65.64 \\
\rowcolor{lightblue}
5 frames       & \textbf{39.77} & \textbf{30.74} & \textbf{67.69} & \textbf{46.07} &\textbf{ 72.46} & \textbf{37.01} & \textbf{90.37} & \textbf{66.61} \\
\hline
\end{tabular}
\vspace{-2mm}
\end{table}

\setlength\tabcolsep{3pt}
\begin{table}[t]
\centering
\caption{Radar point cloud density for 4D radar and LiDAR fusion method (MoRAL).}
\vspace{-1mm}
\label{tab:radar_dense}
\begin{tabular}{c|cccc|cccc}
\hline
\multirow{2}{*}{Radar Frames} & \multicolumn{4}{c|}{All area} & \multicolumn{4}{c}{Driving Corridor} \\
\cline{2-9}
& Car & Ped. & Cyc. & mAP & Car & Ped. & Cyc. & mAP \\
\hline
1 frame   & 69.50 & 67.31 & 76.97 & 71.26 & 89.36 & 76.38 & 91.67 & 85.80 \\
3 frames       & 70.19 & 68.27 & 77.62 & 72.03 & 89.92 & 78.01 & 93.24 & 87.06 \\
\rowcolor{lightblue}
5 frames       & \textbf{71.23}& \textbf{69.67} & \textbf{79.01}& \textbf{73.30}& \textbf{90.91} & \textbf{78.90} & \textbf{96.25} & \textbf{88.68} \\
\hline
\end{tabular}
\vspace{-6mm}
\end{table}
\textbf{Enhancement of radar point cloud density by motion compensation in MRE:} Table \ref{tab:radar_only} and \ref{tab:radar_dense} shows the impact of 4D radar point cloud density on radar-only and 4D radar and LiDAR fusion methods. The radar-only detection backbone is based on PointPillars \cite{pointpillars}. From \ref{tab:radar_only} and \ref{tab:radar_dense}, increasing radar point cloud density with our motion compensation improves detection performance for both single- and multi-modal methods. For radar-only detection, using 5-frame radar leads to 1.95\% mAP improvements in the entire area and 2.00\% in the driving corridor compared to single-frame input. For 4D radar and LiDAR fusion, the mAP increases by 2.04\% and 2.88\% in the two regions.

\textbf{Analysis of motion parameter $\alpha$ in MRE:} 
We also conducted experiments to analyze how the motion parameter $a$ in MRE affects the final detection results. A lower threshold introduces unnecessary compensation of static points, whereas a higher threshold with fewer points moved restricts the overall benefits of motion compensation. As is shown in Table \ref{tab:mos_thresh}, setting  $\alpha =0.5$ achieves the highest mAP. 
\setlength\tabcolsep{5pt}
\begin{table}[t]
\centering
\caption{Analysis of motion parameter $\alpha$.}
\label{tab:mos_thresh}
\vspace{-1mm}
\begin{tabular}{c|cccc|cccc}
\hline
\multirow{2}{*}{$\alpha$} & \multicolumn{4}{c|}{All area} & \multicolumn{4}{c}{Driving Corridor} \\
\cline{2-9}
& Car & Ped. & Cyc. & mAP & Car & Ped. & Cyc. & mAP \\
\hline
0.3       & 70.98 & 69.01 & 78.34 & 72.78 & \textbf{91.07} & 77.82 & 95.47 & 88.12 \\
\rowcolor{lightblue}
0.5       & 71.23 & \textbf{69.67} & \textbf{79.01} & \textbf{73.30} & 90.91 & \textbf{78.90} & \textbf{96.25} & \textbf{88.68} \\
0.7       & \textbf{71.31} & 68.56 & 78.21 & 72.43 & 90.78 & 78.26 & 95.61 & 88.22\\
\hline
\end{tabular}
\vspace{-2mm}
\end{table}

\textbf{Analysis of motion features in MAGF:} Table \ref{tab:mre_feature_analysis} demonstrates the importance of two radar motion features $F_{sa}$ and $F_{fp}$ in MAGF modules. Compared to the baseline, both features, $F_{sa}$ and $F_{fp}$, lead to improved mAP. Notably, using $F_{fp}$ alone achieves the highest AP with 69.71\% and 79.01\% in both areas for pedestrians. The reason lies in the feature propagation in FP layers can better retain fine-grained local details for small and dynamic objects. When both features are utilized, the model achieves the best mAP.
\setlength{\tabcolsep}{4pt}  
\begin{table}[t]
\centering
\caption{Analysis of motion features used in MAGF.}
\vspace{-1mm}
\label{tab:mre_feature_analysis}
\begin{tabular}{cc|cccc|cccc}
\hline
\multicolumn{2}{c|}{{\makecell{Motion\\Features}}}& \multicolumn{4}{c|}{All Area} & \multicolumn{4}{c}{Driving Corridor} \\
\cline{1-2} \cline{3-6} \cline{7-10}
$F_{sa}$ & $F_{fp}$ & Car & Ped. & Cyc. & mAP & Car & Ped. & Cyc. & mAP \\
\hline
  &             & \textbf{71.36} & 69.21& 78.66& 73.08& 90.89& 78.19& 94.01& 87.70\\
\checkmark &     & 71.26 & 69.37 & 78.89 & 73.17 & 90.85 & 78.57 & 94.83 & 88.08 \\
          & \checkmark & 71.20 & \textbf{69.71} & 78.66 & 73.19 & 90.37 & \textbf{79.01} & 95.34 & 88.24 \\
\rowcolor{lightblue}
\checkmark & \checkmark & 71.23& 69.67& \textbf{79.01}& \textbf{73.30}& \textbf{90.91}& 78.90& \textbf{96.25}& \textbf{88.68}\\
\hline
\end{tabular}
\vspace{-5mm}
\end{table}

\section{Discussions}
While compensating moving objects effectively mitigates the "tail" issue, our current 4D radar-based motion compensation assumes absolute radial velocity as the real moving direction and compensates along the radial direction. This becomes less effective for objects moving tangentially, whose radial velocity is zero. A potential solution is to estimate the real moving direction using LiDAR sequences before MRE. Additionally, background mis-segmentation in accumulation may introduce additional noise for static objects. Moving forward, we plan to enhance the reliability of MOS by enforcing temporal motion consistency in point-level predictions.

\section{Conclusion}
In this paper, we propose MoRAL to address the inter-frame misalignment caused by object motion in accumulated 4D radar point clouds and to bridge the density gap between 4D radar and LiDAR data. The MRE module generates motion-compensated point clouds that mitigate the "tail" issue, while the MAGF module selectively enhances LiDAR features using radar motion features to highlight foreground moving objects. By dynamically compensating moving objects and enhancing LiDAR features with radar motion features, our method achieves accurate 3D object detection. Comprehensive experiments on the VoD dataset \cite{palffy2022multi} validate the superiority of our method, particularly in detecting small traffic participants such as pedestrians and cyclists. 

\section*{Acknowledgment}
This research has been conducted as part of the DELPHI project, which is funded by the European Union, under grant agreement No 101104263. 
Views and opinions expressed are those of the author(s) only and do not necessarily reflect those of the European Union or the European Climate, Infrastructure and Environment Executive Agency (CINEA). 
Neither the European Union nor the granting authority can be held responsible for them.



\bibliographystyle{IEEEtran} 
\bibliography{thebib} 
\end{document}